\documentclass[letterpaper]{article} 
\usepackage{aaai2026}  
\usepackage{times}  
\usepackage{helvet}  
\usepackage{courier}  
\usepackage[hyphens]{url}  
\usepackage{graphicx} 
\usepackage{natbib}  
\usepackage{caption} 
\usepackage{newfloat}
\usepackage{listings}
\DeclareCaptionStyle{ruled}{labelfont=normalfont,labelsep=colon,strut=off} 
\usepackage{xcolor,colortbl}
\usepackage[mathscr]{euscript}
 \let\mathscr\relax
\usepackage{amsthm}
\usepackage{amsmath}
\usepackage{amssymb}
\usepackage{calc}
\usepackage{amsfonts}
\usepackage{mathrsfs} 
\usepackage{stmaryrd}
\usepackage{mathtools}
\usepackage{siunitx}
\usepackage{harpoon}
\usepackage{bbold}
\usepackage{wasysym}
\usepackage{multirow}
\usepackage{caption}
\usepackage{subcaption}
\usepackage{diagbox}
\usepackage[shortlabels]{enumitem}
\usepackage{xspace}
\usepackage{etoolbox}
\usepackage{ifthen}
\usepackage{booktabs}
\usepackage{nicematrix}
\usepackage{amssymb}
\usepackage{dsfont}

\newcommand*{\eg}{e.g.\@\xspace}
\newcommand*{\ie}{i.e.\@\xspace}

\DeclareMathOperator*{\argmin}{arg\,min}
\newcommand{\aggregate}{\textsc{Aggregate}\xspace}
\newcommand{\combine}{\textsc{Combine}\xspace}
\newcommand{\best}[1]{\cellcolor{black!20}\bfseries #1}
\newcommand{\multiset}[1]{\{\!\{#1\}\!\}}
\newcommand{\cost}{\mathrm{cost}}
\newcommand{\loss}{\mathcal{L}}
\newcommand{\regularizer}{\Omega\xspace}
\newcommand{\explicitregularizer}{\Omega^{\text{Exp.}}\xspace}
\newcommand{\heuristicregularizer}{\Omega^{\text{Heu.}}\xspace}
\newcommand{\lmcut}{h^{\mathrm{LMcut}}}
\newcommand{\hstar}{h^{\ast}}
\newcommand{\teacheraction}{a^{\ast}}
\newcommand{\nonteacheraction}{a_{i}}
\newcommand{\rgnn}{R-GNN\xspace}
\newcommand{\objectencoding}{OE\xspace}
\newcommand{\objectatomencoding}{OAE\xspace}

\theoremstyle{plain}
\newtheorem{theorem}{Theorem}
\theoremstyle{definition}
\newtheorem{definition}[theorem]{Definition}

\usepackage[linesnumbered,ruled,vlined,noend]{algorithm2e}
\SetAlFnt{\small}
\SetAlCapFnt{\small}
\SetAlCapNameFnt{\small}
\SetArgSty{textnormal}
\SetFuncSty{textnormal}
\SetFuncArgSty{textnormal}
\SetCommentSty{textnormal}
\SetNlSty{bfseries}{\color{black}}{}

\makeatletter
\newcommand{\MySetKwFunction}[2]{%
  \expandafter\gdef\csname @#1\endcsname##1{\FuncSty{#2\ensuremath{(}}\FuncArgSty{##1}\FuncSty{\ensuremath{)}}}%
  \expandafter\gdef\csname#1\endcsname{%
    \@ifnextchar\bgroup{\csname @#1\endcsname}{\FuncSty{#2}\xspace}}%
}
\makeatother
\SetKwInput{Input}{Input}
\SetKwInput{Output}{Output}

\title{Per-Domain Generalizing Policies:\\ On Learning Efficient and Robust Q-Value Functions\\ (Extended Version with Technical Appendix)}
\author{
  Nicola J. Müller\textsuperscript{\rm 1,2,3},
  Moritz Oster\textsuperscript{\rm 1,2},
  Isabel Valera\textsuperscript{\rm 2},
  J\"{o}rg Hoffmann\textsuperscript{\rm 1,\rm 2},
  Timo P. Gros\textsuperscript{\rm 1,2,3}
}
\affiliations {
\textsuperscript{\rm 1}German Research Center for Artificial Intelligence (DFKI), Saarbr\"{u}cken, Germany \\
  \textsuperscript{\rm 2}Saarland University, Saarland Informatics Campus, Saarbr\"{u}cken, Germany \\
  \textsuperscript{\rm 3}Center for European Research in Trusted Artificial Intelligence (CERTAIN) \\
  \{Nicola.Mueller, Moritz.Oster, Timo\_Phillip.Gros\}@dfki.de,
  \{ivalera,hoffmann\}@cs.uni-saarland.de 
}

\begin{document}

\nocopyright

\maketitle


\begin{abstract}
Learning per-domain generalizing policies is a key challenge in learning for planning. 
Standard approaches learn state-value functions represented as graph neural networks using supervised learning on optimal plans generated by a teacher planner.
In this work, we advocate for learning Q-value functions instead.
Such policies are drastically cheaper to evaluate for a given state, as they need to process only the current state rather than every successor. 
Surprisingly, vanilla supervised learning of Q-values performs poorly as it does not learn to distinguish between the actions taken and those not taken by the teacher. 
We address this by using regularization terms that enforce this distinction, resulting in Q-value policies that consistently outperform state-value policies across a range of $10$ domains and are competitive with the planner LAMA-first. 
\end{abstract}

\begin{links}
     \link{Code}{https://doi.org/10.5281/zenodo.19069966}
\end{links}


\section{Introduction}
\label{sec:introduction}
Learning per-domain generalizing policies using neural networks is gaining popularity in planning~\cite{toyer:etal:jair-20,staahlberg2022learning,staahlberg2022blearning,staahlberg2025learning,rossetti2024learning,wang2024learning,chen2025graph,chen2025language}.
The key challenge is to foster scaling behavior, \ie, generalizing from small training instances to large test instances.

Prior work used \emph{graph neural networks} (GNNs) to learn \emph{state-value functions} $V : \mathcal{S} \rightarrow \mathbb{R^{+}}$ for classical planning domains, which approximate the optimal heuristic $h^{*}$~\cite{staahlberg2022learning,staahlberg2022blearning,staahlberg2025learning,gros2025per}.
Given $V$, a per-domain generalizing policy is computed as $\pi(s) = \argmin_{a} \cost(a) + V(s')$, where $s'$ is the respective successor reached through action $a$. 

\begin{figure}[t]
    \centering 
    \includegraphics[width=0.49\linewidth]{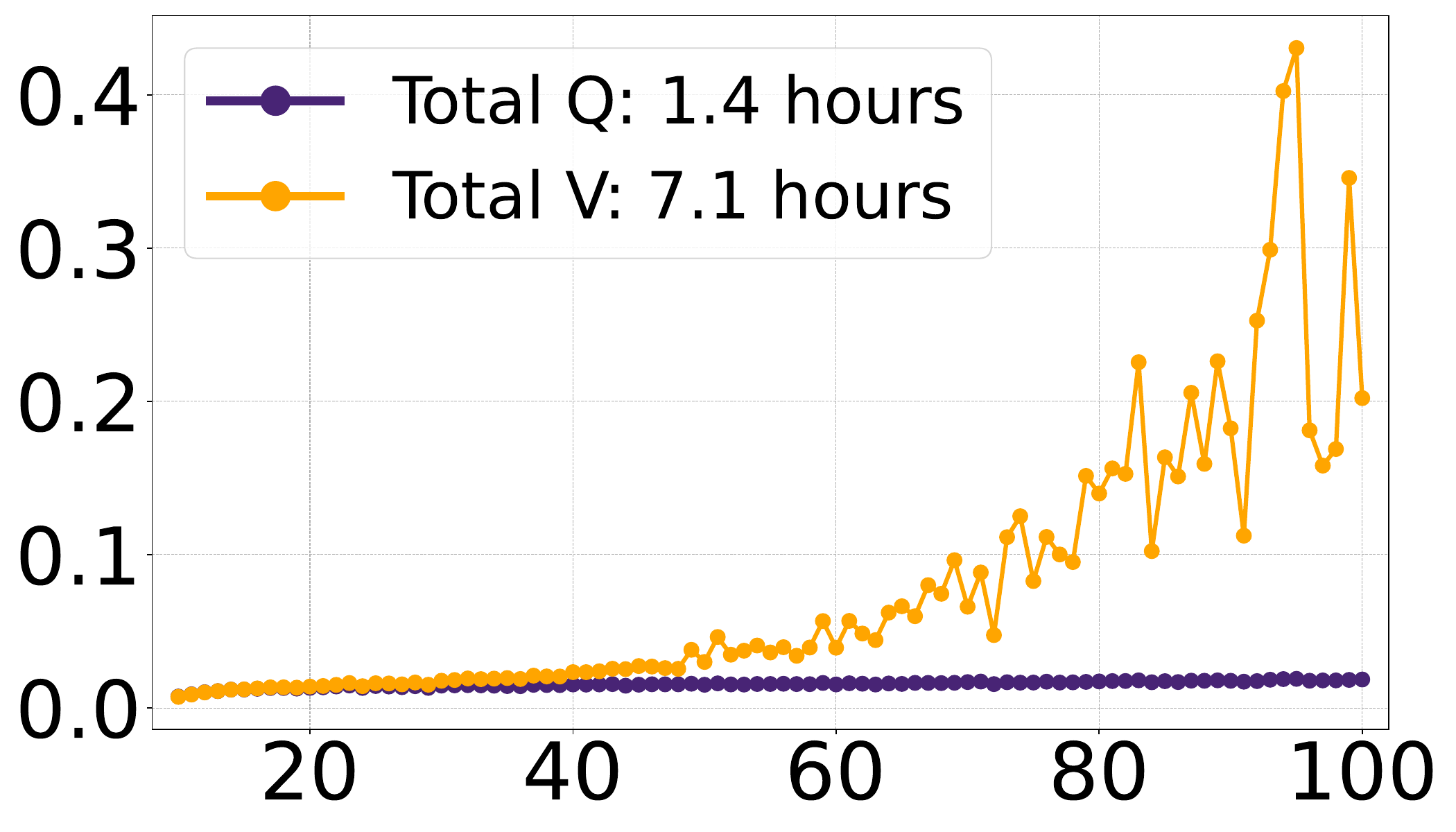}
    \hfill 
    \includegraphics[width=0.49\linewidth]{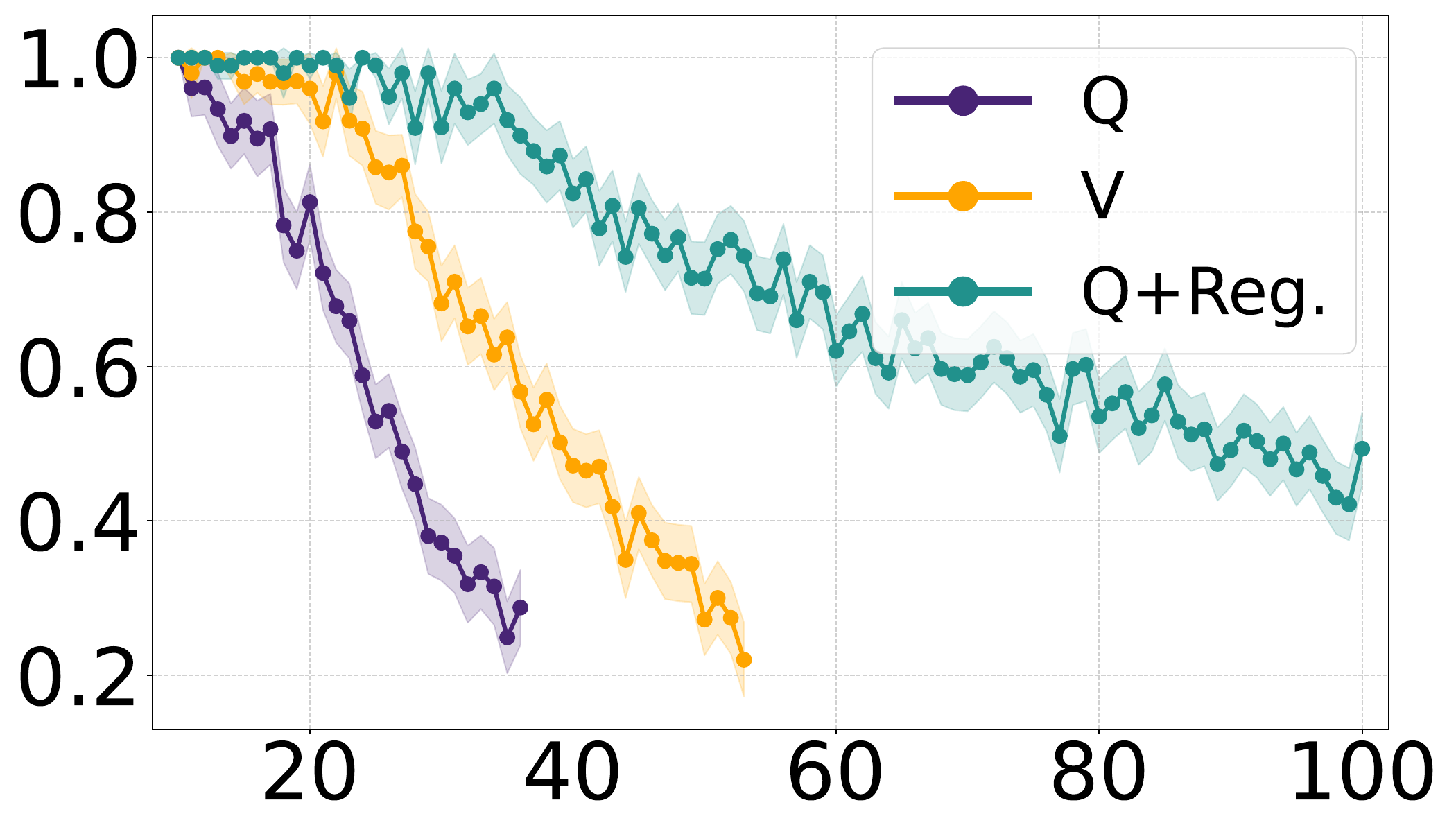}

    \caption{(Left) Total runtime of forward passes in hours per instance size on random states.
    (Right) Average coverage and confidence interval per instance size for trained policies.}
    \label{fig:intro-time-coverage} 
\end{figure}

Observe that choosing an action this way requires the GNN to process every successor state. 
This can be very ineffective and impede scaling performance, even when using batch processing. 
This motivates learning a Q-value function $Q : \mathcal{S} \times Act \rightarrow \mathbb{R^{+}}$ instead, which approximates the cost of an optimal plan starting with a given action $a$ from a given state $s$~\cite{sutton1998reinforcement}.
A policy can then be computed as $\pi(s) = \argmin_{a} Q(s,a)$, which only requires the current state to compute the Q-values of all applicable actions with a single GNN forward pass.
Figure~\ref{fig:intro-time-coverage} (left) shows the runtime per instance size on random states for the IPC'23 domain Rovers~\cite{ayal:etal:aim-23}, where a Q-value policy is $5$ times faster than a state-value policy.

We leverage existing planners to train our Q-value functions using \emph{supervised learning} (SL), where the training data consist of optimal plans~\cite{staahlberg2022learning, rossetti2024learning,gros2025per}.
To the best of our knowledge, this has not been tried before for learning per-domain generalizing policies.
Surprisingly, we show that the vanilla use of SL for Q-values yields policies that generalize poorly, despite fitting the training data well. 
Consider Figure~\ref{fig:intro-time-coverage} (right), which shows the average coverage per instance size of state (orange) vs.\ Q-value (purple) policies.
Both policies were trained on the same optimal plans using the same loss; yet, the state-value policy generalizes much better than the Q-value policy.
This is because the Q-values are often identical for all actions, leading to random action selection. 
In other words, the Q-values fail to distinguish between the actions taken and not taken by the teacher planner.

Our key contribution is showing that this can be fixed.
We extend the SL objective with a regularization term that enforces the Q-values of non-teacher actions to be larger than those of teacher actions.
This approach is shown in green in Figure~\ref{fig:intro-time-coverage} (right), and it vastly outperforms the state-value policy. 
Importantly, this difference is strictly due to better action selection, as policy runs are only limited by the number of actions, without any runtime limit.

We evaluate our approach with three GNN architectures on $10$ domains and find that the trends from Figure~\ref{fig:intro-time-coverage} (right) apply consistently, yielding regularized Q-value policies that outperform state-value policies and are competitive with the planner LAMA-first.


\section{Efficiency of State and Q-Value Policies}
\label{sec:policy_representations}

Here, we show that Q-value policies are more efficient than state-value policies using three \emph{graph neural network} (GNN) architectures for per-domain generalizing policies.

\paragraph{GNN architectures.}
The \emph{relational graph neural network} (\rgnn) is a specialized GNN architecture that operates directly on the relational structure induced by classical planning states, maintaining embeddings of a state's objects and passing information between them according to the ground atoms of which they are arguments~\cite{staahlberg2022learning}.
Conversely, the \emph{object encoding} (\objectencoding) and \emph{object-atom encoding} (\objectatomencoding) define graph representations of states that can be processed by off-the-shelf GNN architectures~\cite{horcik2024expressiveness}; we use the \emph{relational graph convolutional network} (RGCN)~\cite{schlichtkrull2018modeling}, as in prior work~\cite{chen2024learning}.
In a nutshell, \objectencoding defines nodes for a state's objects and adds edges between objects occurring in the same ground atoms, whereas \objectatomencoding defines nodes for both objects and atoms, adding edges between atoms and their arguments.

All three GNN architectures have in common that they iteratively compute embeddings for every object in a given state.
Independent of the architecture, the object embeddings can be aggregated into a state embedding, which is then used to predict either a single state-value or the Q-values of all applicable actions.
We use the approach of \citeauthor{stahlberg2025first-order}~\shortcite{stahlberg2025first-order}. 
For the sake of brevity, we omit the details here and provide a description in the technical appendix~\ref{app:gnns}.

\paragraph{Policy efficiency.}
To compute a policy using state-values, \ie, $\pi(s) = \argmin_{a} \cost(a) + V(s')$, we need to predict the values of all successor states $s'$.
This requires expanding the current state, translating the successors into GNN input, combining them into a single batch, and passing it to the GNN.
To compute a policy using Q-values, \ie, $\pi(s) = \argmin_{a} Q(s,a)$, we need to predict the values of all applicable actions in the current state $s$.
This only requires translating the current state into GNN input and passing it to the GNN.
Hence, state-value policies are less efficient than Q-value policies because they require processing every successor of the current state instead of the current state only.
When scaling the instance size, the overhead for computing state-values grows drastically, as a larger instance size typically coincides with a larger branching factor, and, thus, more successors to process at every state.

Consider Table~\ref{tab:efficiency}, which compares the runtimes of state and Q-value policies on the $10$ domains used in Section~\ref{sec:experiments}.
For each domain, we generate a set of states using random walks of length $25$ for $25$ instances per size, ranging from the smallest possible instance size up to $100$.
For each state, we predict an action using a randomly initialized state or Q-value policy and record the total runtime.
All policies are executed on an NVIDIA RTX A$6000$ GPU.
Across all architectures, we see that the average runtimes of the state-value policies are between $4$ and $18$ times higher than those of the Q-value policies.
We conclude that Q-values should be preferred over state-values for per-domain generalizing policies, as they scale much more efficiently with instance size.

\begin{table}
  \centering
  \setlength{\tabcolsep}{5pt}    
  \renewcommand{\arraystretch}{0.7}
  \scalebox{0.9}{
    \begin{tabular}{l *{2}{c} *{2}{c} *{2}{c}}
        & \multicolumn{2}{c}{\rgnn} & \multicolumn{2}{c}{\objectencoding} & \multicolumn{2}{c}{\objectatomencoding} \\
        \cmidrule(lr){2-3} \cmidrule(lr){4-5} \cmidrule(lr){6-7}
        Domain & $V$ & $Q$ & $V$ & $Q$ & $V$ & $Q$ \\
        \midrule
		Blocksworld & 0.6 & 0.5 & 1.8 & 2.0 & 0.7 & 0.4 \\ 

		Childsnack & 7.2 & 0.9 & 21.7 & 4.3 & 28.6 & 0.7 \\ 

		Ferry & 0.8 & 0.5 & 2.5 & 2.0 & 1.6 & 0.4 \\ 

		Floortile & 1.9 & 1.0 & 7.9 & 4.0 & 4.0 & 0.6 \\ 

		Gripper & 1.0 & 0.6 & 3.1 & 2.2 & 1.9 & 0.5 \\ 

		Logistics & 2.7 & 0.7 & 7.3 & 2.3 & 10.2 & 0.6 \\ 

		Rovers & 7.1 & 1.4 & 39.8 & 7.5 & 25.4 & 1.1 \\ 

		Satellite & 4.6 & 0.8 & 14.7 & 3.5 & 16.0 & 0.7 \\ 

		Transport & 6.5 & 0.7 & 22.8 & 2.7 & 18.0 & 0.7 \\ 

		Visitall & 0.4 & 0.2 & 1.1 & 0.6 & 0.9 & 0.3 \\ 

		\midrule
		Average & 3.3 & 0.7 & 12.3 & 3.1 & 10.7 & 0.6 \\ 
    \end{tabular}
  }
  \caption{Total runtime in hours for randomly initialized state and Q-value policies on sets of random states.}
  \label{tab:efficiency}
\end{table}


\section{Regularization Terms for Supervised Learning of Q-Value Functions}
\label{sec:regularizers}

We now show why vanilla supervised learning (SL) fails to learn Q-value functions for per-domain generalizing policies, and then fix this by introducing regularization terms.

\subsection{Supervised Learning of Q-Value Functions}
Prior work learned state-value functions using SL, where the training data consist of optimal plans for small domain instances~\cite{staahlberg2022learning,gros2025per}.
However, we find that learning Q-value functions using this approach yields policies that fail to generalize.
This is because the Q-value functions tend to predict the same Q-values for all actions applicable in a given state, leading to random action selection.

We investigate this behavior by comparing Q-value functions successfully learned for $10$ domains using the \rgnn, \objectencoding, and \objectatomencoding architectures.
We refer to Section~\ref{sec:experiments} for training details.
For each model, we compute the average prediction error for the Q-values of teacher actions $\teacheraction$ in the training set ($\mathrm{Err}$), \ie, $\left\vert\hstar(s)-Q(s,\teacheraction)\right\vert$.
We also compute the average difference in predicted Q-values between teacher actions $\teacheraction$ and non-teacher actions $\nonteacheraction \in Act(s) \backslash \{\teacheraction\}$ in the training set ($\mathrm{Diff}$), \ie, $\left\vert Q(s,\teacheraction)-Q(s,\nonteacheraction)\right\vert$.
We present the averages of both scores over $10$ domains.
Consider the left column ($Q$) in Table~\ref{tab:q-differences}.
All models fit their training data well, as the average prediction errors for teacher actions ($\mathrm{Err}$) are small, with values between $0.54$ and $2.15$.
However, the models tend to predict the same Q-values for teacher and non-teacher actions, as the average Q-value differences ($\mathrm{Diff}$) are between $0.71$ and $0.96$.
Hence, Q-value policies learned using vanilla SL perform poorly because they fail to generalize to non-teacher actions.

\begin{table*}[t]
  \centering
  \setlength{\tabcolsep}{5pt}
  \renewcommand{\arraystretch}{0.7}
  \scalebox{0.9}{
    \begin{tabular}{l *{6}{c} *{6}{c} *{6}{c}}
        & \multicolumn{6}{c}{$Q$} & \multicolumn{6}{c}{$\explicitregularizer$} & \multicolumn{6}{c}{$\heuristicregularizer$} \\
        \cmidrule(lr){2-7} \cmidrule(lr){8-13} \cmidrule(lr){14-19}
        & \multicolumn{2}{c}{\rgnn} & \multicolumn{2}{c}{\objectencoding} & \multicolumn{2}{c}{\objectatomencoding}
        & \multicolumn{2}{c}{\rgnn} & \multicolumn{2}{c}{\objectencoding} & \multicolumn{2}{c}{\objectatomencoding}
        & \multicolumn{2}{c}{\rgnn} & \multicolumn{2}{c}{\objectencoding} & \multicolumn{2}{c}{\objectatomencoding}\\
        
        \cmidrule(lr){2-3} \cmidrule(lr){4-5} \cmidrule(lr){6-7} 
        \cmidrule(lr){8-9} \cmidrule(lr){10-11} \cmidrule(lr){12-13} 
        \cmidrule(lr){14-15}\cmidrule(lr){16-17}\cmidrule(lr){18-19} 

        & Err & Diff & Err & Diff & Err & Diff 
        & Err & Diff & Err & Diff & Err & Diff 
        & Err & Diff & Err & Diff & Err & Diff\\
        \cmidrule(lr){2-7} \cmidrule(lr){8-13} \cmidrule(lr){14-19}
        
        Average & 0.54 & 0.71 & 1.37 & 0.96 & 2.15 & 0.84 & 0.70 & 4.69 & 0.70 & 7.66 & 0.91 & 6.66 & 0.73 & 97.55 & 1.25 & 46.25 & 0.83 & 152.52 \\
    \end{tabular}
    }
  \caption{Prediction errors for teacher actions ($\mathrm{Err}$), and Q-value differences between teacher and non-teacher actions ($\mathrm{Diff}$).}
  \label{tab:q-differences}
\end{table*}

\subsection{Regularizing Q-Value Functions}
To ensure that learned Q-value functions distinguish between teacher and non-teacher actions, we extend the SL objective with a regularization term.
The extended learning objective is defined as
\begin{equation*}
    \theta^{\ast} = \argmin_{\theta} \mathds{E}_{(s,\teacheraction) \sim D} \left[ \loss(s, \teacheraction) + \lambda \cdot \regularizer(s, \teacheraction) \right],
\end{equation*}
where $\theta^{\ast}$ are the parameters of the Q-value function $Q$ that minimizes the loss $\loss$ subject to the regularizer $\regularizer$, and $\lambda$ is the regularization coefficient.
The role of $\loss$ is to ensure that $Q$ predicts the Q-values of teacher actions $\teacheraction$ accurately, whereas $\regularizer$ ensures that the Q-values of non-teacher actions $\nonteacheraction$ are larger than those of teacher actions $\teacheraction$, \ie, $\forall \nonteacheraction \in Act(s) \backslash \{\teacheraction\} : Q(s, \nonteacheraction) > Q(s, \teacheraction)$.
We define the regularization term in general as
\begin{equation*}
	\regularizer(s, \teacheraction) = \sum_{\nonteacheraction \in Act(s) \backslash \{\teacheraction\}}\max \left\{0, B - Q(s, \nonteacheraction) \right\},
\end{equation*}
where $B$ is a lower bound for the Q-values of non-teacher actions $\nonteacheraction$.
The penalty induced by $\regularizer$ increases linearly if $Q(s, \nonteacheraction)$ is smaller than $B$, whereas there is no penalty if $Q(s, \nonteacheraction) \ge B$ holds.
We now introduce two regularizers $\regularizer$ that differ in how they compute the lower bound $B$.

\paragraph{Explicit regularizer.}
The \emph{explicit regularizer} $\explicitregularizer$ computes $B$ using the $\hstar$ values of the states $s$ in the training data, which is feasible as they are on optimal trajectories.
We define the lower bound as
\begin{equation*}
	B^{\text{Exp.}}(s) = \hstar(s) + 1,
\end{equation*}
ensuring that the Q-values of non-teacher actions $Q(s, \nonteacheraction)$ are strictly larger than the target values of teacher actions $Q(s,\teacheraction) = \hstar(s)$. 
While this biases learning towards an inadmissible Q-value function, as there might be multiple optimal actions, we can still converge to an optimal policy since teacher actions always have the lowest Q-values.
We note that, in general, admissibility cannot be guaranteed for a learned Q-value function, even without this bias.

\paragraph{Heuristic regularizer.}
The \emph{heuristic regularizer} $\heuristicregularizer$ uses an admissible heuristic to compute \emph{action-dependent} lower bounds $B_{i}$, which provide more information about the true Q-values of non-teacher actions $\nonteacheraction$.
First, we compute a lower bound $\cost(\nonteacheraction) + h(s'_{i})$ for every $\nonteacheraction$, where $s'_{i}$ is the successor reached through action $\nonteacheraction$, and $h$ is an admissible heuristic; in this paper, $\lmcut$~\cite{helmert2009landmarks}\footnote{For dead-ends, \ie, $\lmcut(s'_{i})=\infty$, we use a value of $1120$, which is drastically larger than all plans in our training data.}.
However, this lower bound may not be strictly larger than the target Q-value of the teacher action, \ie, $\hstar(s)$, preventing us from distinguishing between non-teacher and teacher actions.
Thus, we compute the final lower bound as
\begin{equation*}
	B^{\text{Heu.}}(s,\nonteacheraction) = \max\left\{ \hstar(s)+1, \cost(\nonteacheraction)+h(s'_{i}) \right\}.
\end{equation*}
On average, $B^{\text{Heu.}}$ is tighter than $B^{\text{Exp.}}$ for $17\%$ of the non-teacher actions in the training sets used in Section~\ref{sec:experiments}. 

\paragraph{Comparing regularizers.}
Consider again Table~\ref{tab:q-differences}, where the middle ($\explicitregularizer$) and right ($\heuristicregularizer$) columns show the prediction errors ($\mathrm{Err}$) and Q-value differences ($\mathrm{Diff}$) for Q-value functions trained using the explicit and heuristic regularizers, respectively.
Compared to vanilla SL on the left ($Q$), the Q-value differences increase while the prediction errors mostly decrease, suggesting that the regularizers not only help to distinguish between non-teacher and teacher actions but can also help to predict the Q-values of teacher actions more accurately.
The $\heuristicregularizer$ models also have considerably larger Q-value differences than the $\explicitregularizer$ models, which is due to the high predicted value for actions leading to dead-ends.


\section{Experiments}
\label{sec:experiments}

\begin{table*}[t]
  \centering
  \setlength{\tabcolsep}{2pt}    
  \renewcommand{\arraystretch}{1.1}
  \scalebox{0.71}{
    \begin{tabular}{l *{8}{c} *{8}{c} *{8}{c}}
        & \multicolumn{8}{c}{\rgnn} & \multicolumn{8}{c}{\objectencoding} & \multicolumn{8}{c}{\objectatomencoding} \\
        \cmidrule(lr){2-9} \cmidrule(lr){10-17} \cmidrule(lr){18-25}
        & \multicolumn{2}{c}{$V$} & \multicolumn{2}{c}{$Q$} & \multicolumn{2}{c}{$\explicitregularizer$} & \multicolumn{2}{c}{$\heuristicregularizer$}
        & \multicolumn{2}{c}{$V$} & \multicolumn{2}{c}{$Q$} & \multicolumn{2}{c}{$\explicitregularizer$} & \multicolumn{2}{c}{$\heuristicregularizer$}
        & \multicolumn{2}{c}{$V$} & \multicolumn{2}{c}{$Q$} & \multicolumn{2}{c}{$\explicitregularizer$} & \multicolumn{2}{c}{$\heuristicregularizer$}\\
        \cmidrule(lr){2-3}  \cmidrule(lr){4-5}  \cmidrule(lr){6-7} \cmidrule(lr){8-9}
        \cmidrule(lr){10-11}  \cmidrule(lr){12-13} \cmidrule(lr){14-15} \cmidrule(lr){16-17}
        \cmidrule(lr){18-19}\cmidrule(lr){20-21}\cmidrule(lr){22-23} \cmidrule(lr){24-25}
        Domain & Scale & SCov & Scale & SCov & Scale & SCov & Scale & SCov 
        & Scale & SCov & Scale & SCov & Scale & SCov & Scale & SCov
        & Scale & SCov & Scale & SCov & Scale & SCov & Scale & SCov\\
        \midrule
        
		Blocksworld & 52 & 44.5 & 5 & 1.6 & 54 & 47.8 & \best{55} & \best{48.2} & 14 & 9.2 & 5 & 1.2 & 48 & 40.5 & \textbf{50} & \textbf{42.4} & 37 & 27.5 & 4 & 1.3 & \textbf{52} & \textbf{44.9} & \textbf{52} & 44.6 \\

		Childsnack & 71 & 43.4 & 16 & 4.2 & \best{100} & \best{93.0} & \best{100} & \best{93.0} & 26 & 12.0 & 9 & 0.2 & 82 & 43.1 & \textbf{98} & \textbf{46.9} & 31 & 15.9 & 36 & 14.9 & \best{100} & \textbf{65.8} & 83 & 41.5 \\

		Ferry & \best{100} & 84.5 & 30 & 12.9 & \best{100} & \best{85.2} & \best{100} & 84.5 & 33 & 19.8 & 21 & 10.9 & \best{100} & \textbf{80.2} & \best{100} & 80.1 & 45 & 26.4 & 7 & 2.1 & \best{100} & 79.5 & \best{100} & \textbf{80.1} \\

		Floortile & 8 & 0 & 8 & 0 & 60 & 44.5 & \best{62} & \textbf{45.6} & 8 & 0 & 8 & 0 & 50 & 28.8 & \textbf{56} & \textbf{43.2} & 8 & 0 & 8 & 0 & 59 & 42.0 & \best{62} & \best{45.9} \\

		Rovers & 53 & 29.8 & 36 & 17.1 & \best{100} & \best{66.5} & \best{100} & 64.4 & 25 & 9.8 & 33 & 13.3 & \textbf{48} & \textbf{24.6} & 46 & 23.0 & 22 & 8.3 & 28 & 9.8 & 60 & \textbf{30.1} & \textbf{61} & 29.8 \\

		Satellite & 50 & 28.2 & 37 & 21.5 & \best{100} & \best{69.4} & \best{100} & 69.1 & 20 & 9.8 & 25 & 12.9 & \textbf{73} & \textbf{45.3} & \textbf{73} & 45.2 & 24 & 9.7 & 6 & 0.2 & \textbf{92} & 55.1 & 90 & \textbf{55.5} \\

		Transport & 72 & 46.6 & 46 & 24.7 & \best{100} & 90.4 & \best{100} & \best{92.7} & 38 & 20.1 & 42 & 23.1 & \best{100} & 88.6 & \best{100} & \textbf{89.7} & 19 & 9.3 & 57 & 32.1 & \best{100} & \textbf{89.7} & \best{100} & \textbf{89.6} \\

		Gripper & \best{100} & 77.5 & 36 & 15.2 & \best{100} & 88.8 & \best{100} & \best{88.9} & \best{100} & 83.0 & 63 & 33.4 & \best{100} & \best{88.9} & \best{100} & 88.4 & \best{100} & \textbf{88.6} & \best{100} & 73.8 & \best{100} & 88.5 & \best{100} & \textbf{88.6} \\

		Logistics & 47 & 24.9 & 50 & 25.9 & \best{100} & 82.4 & \best{100} & \textbf{84.3} & 36 & 19.8 & 52 & 27.0 & \best{100} & \best{86.2} & \best{100} & 85.7 & 38 & 22.6 & 57 & 31.2 & \best{100} & \textbf{77.0} & \best{100} & 75.6 \\

		Visitall & 256 & 11.4 & 256 & 11.4 & \best{961} & \best{25.1} & \best{961} & 22.8 & 196 & 9.5 & 144 & 5.7 & \best{961} & \textbf{22.7} & \best{961} & 22.1 & 529 & 13.4 & 484 & 14.9 & \best{961} & \textbf{24.8} & 676 & 16.6 \\ 

        \midrule
    \end{tabular}
  }
  \caption{Scale and SCov scores of the scaling behavior evaluation. 
  Bold scores are the best for each architecture, and highlighted scores are the best overall. 
  }
  \label{tab:scaling_behavior}
\end{table*}

\begin{table}[t]
  \centering
  \setlength{\tabcolsep}{4pt}   
  \renewcommand{\arraystretch}{1.0}
  \scalebox{0.8}{
    \begin{tabular}{l *{6}{c} *{2}{c}}
        & \multicolumn{6}{c}{\rgnn} \\
        \cmidrule(lr){2-7}
        & \multicolumn{2}{c}{$V$} & \multicolumn{2}{c}{$\explicitregularizer$} & \multicolumn{2}{c}{$\heuristicregularizer$}
        & \multicolumn{2}{c}{LAMA-first}\\
        \cmidrule(lr){2-3}  \cmidrule(lr){4-5}  \cmidrule(lr){6-7}
        \cmidrule(lr){8-9}
        Domain & Cov & Len & Cov & Len & Cov & Len & Cov & Len\\
        \midrule
		Blocksworld & 47.8 & 90.0 & \textbf{78.9} & 233.1 & \textbf{78.9} & 232.1 & 56.7 & 288.7 \\ 
		Childsnack & 5.6 & 41.2 & 51.1 & 41.4 & \textbf{54.4} & 41.5 & 35.2 & 41.8 \\ 
		Ferry & 54.4 & 73.7 & \textbf{67.8} & 119.6 & \textbf{67.8} & 119.6 & \textbf{67.8} & 150.3 \\ 
		Floortile & 0.0 & --- & 36.7 & 93.5 & \textbf{37.8} & 109.3 & 16.9 & 51.0 \\ 
		Rovers & 24.4 & 49.0 & 36.7 & 66.2 & 37.8 & 80.3 & \textbf{57.8} & 171.0 \\ 
		Satellite & 32.2 & 36.9 & 64.4 & 71.2 & 66.7 & 68.2 & \textbf{85.6} & 104.4 \\ 
		Transport & 24.4 & 69.6 & 70.0 & 91.3 & \textbf{74.4} & 96.8 & 64.4 & 70.7 \\ 
		 \midrule
		\textbf{Average} & 27.0 & 60.1 & 57.9 & 102.3 & \textbf{59.7} & 106.8 & 54.9 & 125.4 \\ 
        \midrule
    \end{tabular}
  }
  \caption{Coverage in percent (Cov) and average plan length (Len) on IPC'23 test sets. Bold scores are the best coverages for each domain.}
  \label{tab:ipc_results}
\end{table}

In this section, we outline our data set construction and training setup, and then evaluate our approaches with respect to scaling behavior and IPC test set performance.

\paragraph{Data sets.}
We consider the domains from the IPC'23 learning track~\cite{ayal:etal:aim-23}, where we omit Miconic and Spanner because all policies immediately achieve $100\%$ coverage, and Sokoban because its generator requires unfeasibly many runs until a solvable instance is generated.
We also consider the Gripper, Logistics, and Visitall domains from the FF domain collection\footnote{\url{https://fai.cs.uni-saarland.de/hoffmann/ff-domains.html}}.

To construct training sets for each domain, we use the publicly available generators to uniformly sample up to $100$ unique instances per size, ranging from size $2$ to $100$.
We use the \textit{seq-opt-merge-and-shrink} configuration of Fast Downward~\cite{helmert:jair-06} with limits of $20$ minutes and $64$ GB as the optimal teacher planner, deleting instances that were not solved.
We provide full details about the training sets in the technical appendix~\ref{app:training}.
For validation, we use \citeauthor{gros2025per}'s dynamic coverage validation, which does not require a precomputed validation set, as it runs policies on increasingly large instances generated on-the-fly~\shortcite{gros2025per}.

\paragraph{Training setup.}
We train state, Q, and regularized Q-value functions using the \rgnn, \objectencoding, and \objectatomencoding architectures.
As the loss function $\loss$, we use the \emph{mean absolute error} ($\mathrm{MAE}$), which is defined as $\mathrm{MAE}(s)=\left \vert \hstar(s)-V(s)\right \vert$ for state-values and as $\mathrm{MAE}(s, \teacheraction)=\left \vert \hstar(s)-Q(s,\teacheraction)\right \vert$ for Q-values.
In preliminary experiments, we also tested the mean squared error and temporal difference error losses, but found that $\mathrm{MAE}$ yields the best policies.
For the regularized Q-value functions, we use a regularization coefficient of $\lambda=1.0$, giving equal weight to $\loss$ and $\regularizer$.
We repeat the training using three random seeds and return the policy with the best validation performance.
When running the policies, we forbid revisiting states to prevent cycles, as in prior work~\cite{staahlberg2022learning, staahlberg2022blearning, staahlberg2023learning}.

\paragraph{Scaling behavior evaluation.}
We use \citeauthor{gros2025per}'s scaling behavior evaluation to compare how well the trained policies generalize~\shortcite{gros2025per}.
For a given instance size $n$, the evaluation runs a policy on uniformly generated instances until the estimate of the average coverage $\hat{C}_{n}$ for size $n$ is within a $10\%$ confidence interval with $90\%$ probability.
This process repeats with an increasing size $n$ until the coverage $\hat{C}_{n}$ drops below $30\%$ or the maximum size of $100$ ($1000$ for Visitall) is reached.
Policy runs have a step limit of $100 + n$ that increases with the instance size $n$.
The evaluation results are reported using two summary scores: \emph{Scale} is the largest instance size $n$ before termination, which measures how far the policy generalizes, and \emph{SCov} is the sum of coverages $\hat{C}_{n}$ per size $n$, which measures overall performance.

Table~\ref{tab:scaling_behavior} shows the summary scores for the scaling behavior evaluation of all policies.
We observe the following trends across all architectures:
The vanilla Q-value policies ($Q$) perform the worst overall.
This is due to failing to distinguish teacher and non-teacher actions, as shown in Section~\ref{sec:regularizers}.
Importantly, we see that the state-value policies ($V$) are consistently outperformed by the regularized Q-value policies ($\explicitregularizer$ \& $\heuristicregularizer$).
This shows that the additional learning signals provided by regularizing the Q-values of non-teacher actions lead to vastly improved generalization.
Lastly, we see that the policies trained using the heuristic regularizer ($\heuristicregularizer$) outperform the explicit regularizer policies ($\explicitregularizer$) on some domains.
This shows that providing tighter lower bounds for the Q-values of non-teacher actions can further improve generalization.

\paragraph{IPC test sets.}
We additionally evaluate our policies on the IPC'23 test sets, which provides insights into their performance on particularly difficult instances.
We consider only the \rgnn architecture, as it showed the overall best performance in the scaling behavior evaluation, and we omit the vanilla Q-value policies due to their poor performance.
We ensured that the training and test sets do not overlap.
All policies are executed on an AMD EPYC $9454$ CPU with time and memory limits of $1$ minute and $8$ GB.
As a baseline, we run the LAMA-first planner~\cite{richter2010lama} using the same hardware and limits.

Consider Table~\ref{tab:ipc_results}, which shows the coverage and average plan length for the $7$ considered IPC'23 domains.
As before, our regularized Q-value policies consistently achieve higher coverages than the state-value policies.
Impressively, the coverages and plan lengths of the regularized Q-value policies are competitive with those of LAMA-first.


\section{Conclusion}
\label{sec:conclusion_and_future_work}

Learning per-domain generalizing policies is becoming increasingly popular, and supervised learning provides a natural training framework as we can leverage optimal planners to generate rich training data.
In this work, we have shown that the common approach of learning state-value functions yields policies whose runtime scales poorly with instance size.
The more efficient approach of learning Q-value functions, however, fails to learn the distinction between teacher and non-teacher actions.
Hence, we introduce regularization terms that use optimal and admissible heuristics to enforce the Q-values of non-teacher actions to be larger than those of teacher actions.
Experiments on $10$ domains using three GNN architectures show that our regularized Q-value policies are not only more efficient but also achieve vastly higher generalization than state-value policies.

A promising next step for our work would be to continue the policy training using reinforcement learning on dynamically generated instances.
These instances could be generated as needed, \eg, we could gradually increase their size to improve scaling or focus on instances where the policy previously failed.


\section*{Acknowledgements}
This work was partially supported by the German Federal Ministry of Education and Research (BMBF) as part of the project MAC-MERLin (Grant Agreement No. 01IW24007), by the German Research Foundation (DFG)  - GRK 2853/1 “Neuroexplicit Models of Language, Vision, and Action” - project number 471607914, and by the European Regional Development Fund (ERDF) and the Saarland within the scope of (To)CERTAIN.

\bibliography{aaai2026}


\newpage
\appendix
\section{Classical Planning}
\label{app:background}
A classical planning problem can be represented as a pair $\langle D,I \rangle$, consisting of a domain $D$ and an instance $I$~\cite{ghallab2004automated}. 
The domain $D$ defines a set of predicates $P \in \mathcal{P}$ and a set of action schemas $A \in \mathcal{A}$, each of which describes arguments, preconditions, effects, and action costs.
The instance $I$ defines a set of objects $o \in \mathcal{O}$, using which we can derive ground atoms $p := P(o_{0}, \dots, o_{n})$ and ground actions $a := A(o_{0}, \dots, o_{n})$.
Further, $I$ defines an initial state $Init$ and a set of goal conditions $\mathcal{G}$, both of which are sets of ground atoms.
Together, $D$ and $I$ encode a state model $\langle \mathcal{S}, s_{0}, \mathcal{S}_{\mathcal{G}}, Act, f \rangle$, consisting of a set of states $s \in \mathcal{S}$, the initial state $s_{0}$, the set of goal states $\mathcal{S}_{\mathcal{G}}$, the sets of applicable ground actions for each state $Act(s)$, and the transition function $f: \mathcal{S} \times Act \rightarrow \mathcal{S}$.

A plan is a sequence of ground actions $\vec{a} = \langle a_{0}, \dots, a_{T-1} \rangle$ that transitions from the initial state $s_{0}$ to any goal state $s_{T} \in \mathcal{S}_{\mathcal{G}}$.
A plan $\vec{a}^{*}$ is optimal if the sum of the costs of its actions is minimal.
The optimal heuristic value $h^{\ast}(s)$ of a state $s$ is defined as the cost of an optimal plan starting in state $s$.

\section{Graph Neural Networks for Per-Domain Generalizing Policies}
\label{app:gnns}
Computing a per-domain generalizing policy requires a neural network architecture that handles variable-sized inputs because the number of ground atoms that can be true in any given state increases with the instance size.
\emph{Graph neural networks} (GNNs) are one such architecture, and they have become a standard choice for learning per-domain generalizing policies because classical planning states can be effectively represented as graphs~\cite{staahlberg2022learning, chen2024learning,chen2025graph,horvcik2025state}.

\paragraph{Graph Neural Networks.}
Given a graph $G = \langle \mathcal{V}, \mathcal{E} \rangle$ with nodes $\mathcal{V}$ and edges $\mathcal{E}$, a GNN computes for every node $v \in \mathcal{V}$ an embedding $h_{v}$.
Each embedding $h_{v}^{l}$ is iteratively computed over $L$ GNN layers $l$, where the initial embedding $h_{v}^{0}$ corresponds to the node's feature vector as specified by $G$. 
The computation of each layer can be divided into two steps:
\begin{enumerate}
    \item \aggregate: The embeddings $h_{u}^{l}$ of all nodes $u$ in the neighborhood $\mathcal{N}(v)$, i.e.,  $\forall u \in \mathcal{V}: \langle u,v \rangle \in \mathcal{E}$, are aggregated.
    \item \combine: The aggregated embeddings $h_{u}^{l}$ and the node's current embedding $h_{v}^{l}$ are combined to compute the updated embedding $h_{v}^{l+1}$.
\end{enumerate}
These steps can be expressed as the general update rule
\begin{equation}
\label{eq:gnn}
    h_{v}^{l+1} = \combine(h_{v}^{l}, \aggregate(\multiset{h_{u}^{l} \: \vert \: u \in \mathcal{N}(v)})), 
\end{equation}
where $\multiset{\dots}$ denotes a multiset. The update is applied to each node $v \in \mathcal{V}$ simultaneously and for each layer $l$ iteratively until we obtain the final node embeddings $h_{v}^{L}$.  

To demonstrate the generality of our findings and methods, we consider three different GNN architectures for learning per-domain generalizing policies.

\paragraph{Object \& object-atom encodings.}
The \emph{object encoding} (\objectencoding) and \emph{object-atom encoding} (\objectatomencoding) define graph representations of classical planning states that can be processed by off-the-shelf GNN architectures~\cite{horcik2024expressiveness, horvcik2025state}.
The key difference between these encodings is that \objectencoding defines nodes for a state's objects, whereas \objectatomencoding defines nodes for objects and ground atoms.

To encode information about the goal conditions $\mathcal{G}$, we introduce a goal version $P_{\mathcal{G}}$ of each predicate $P \in \mathcal{P}$ and extend every state $s\in \mathcal{S}$ with the corresponding ground atoms $P_{\mathcal{G}}(o_{0}, \dots, o_{n}) \in \mathcal{G}$.
Further, we assign a unique identifier $i \in \mathbb{N}$ to every predicate $P_{i} \in \mathcal{P}$.
Lastly, we treat nullary atoms as unary atoms that hold for all objects, \ie, $\forall o \in \mathcal{O}: P_{i}(o) \in s$.

\definition[Object Encoding]{
For a given state $s$, \objectencoding defines its graph representation as
$G=\left\langle \mathcal{V}, \mathcal{E}, f , l \right\rangle$, 
where
\begin{itemize}
    \item $\mathcal{V} = \mathcal{O}$
    \item $\mathcal{E} = \left\{\langle o_{u},o_{w}\rangle_{P_{i}} \: \middle \vert \: P_{i}(\dots,o_{u},\dots,o_{w},\dots) \in s \right\}$
    \item $f : \mathcal{V} \rightarrow \left\{0,1\right\}^{m}$, with $f(o)_{i} = \mathds{1}\left[ P_{i}(o)\in s \right]$
    \item $l : \mathcal{E} \rightarrow \mathbb{N}$, with $l(\langle o_{u},o_{w} \rangle_{P_{i}}) = i$.
\end{itemize}
In other words, we have a node for every object $o$ with a node feature $f(o)$ that is a multi-hot encoding of the unary predicates $P_{i}$ for which there is an atom $P_{i}(o) \in s$.
Further, if two objects $o_{u}$ and $o_{w}$ are arguments of an atom $P_{i}(\dots,o_{u},\dots,o_{w},\dots) \in s$, we have an undirected edge $\left\langle o_{u},o_{w}\right\rangle_{P_{i}}$ with a label $l(\left\langle o_{u},o_{w}\right\rangle_{P_{i}})$ corresponding to the predicate's identifier $i$.
}

\definition[Object-Atom Encoding]{
For a given state $s$, \objectatomencoding defines its graph representation as
$G=\left\langle \mathcal{V}, \mathcal{E}, f , l \right\rangle$, 
where
\begin{itemize}
    \item $\mathcal{V} = \mathcal{O} \cup s$
    \item $\mathcal{E} = \left\{\langle o_{u}, p \rangle \: \vert \: p:=P_{i}(\dots,o_{u},\dots) \in s\right\}$
    \item $f : \mathcal{V} \rightarrow \left\{0,1\right\}^{m}$, with $f(o)_{i} = \mathds{1}\left[ P_{i}(o)\in s \right]$ and \\ $f(p)_{i} = \mathds{1}\left[ p = P_{i}(\dots) \right]$
    \item $l : \mathcal{E} \rightarrow \mathbb{N}$, with $l(\langle o_{u}, p \rangle) = u$.
\end{itemize}
In other words, we have a node for every object $o$ and ground atom $p$, where objects' node features $f(o)$ are multi-hot encodings of the unary atoms $P_{i}(o) \in s$, and atoms' node features $f(p)$ are one-hot encodings of their predicate $P_{i}$.
We connect atoms $p$ to their arguments $o_{u}$ via undirected edges $\langle o_{u}, p\rangle$, which are labeled by the argument's position $u$.
}

We process the \objectencoding and \objectatomencoding graphs using the \emph{relational graph convolutional network} (RGCN)~\cite{schlichtkrull2018modeling} architecture with the embedding update rule
\begin{equation*}
     h_{v}^{l+1} = \sigma \left ( W^{l}h_{v}^{l} + \sum_{r\in R} \max_{u\in \mathcal{N}_{r}(v)}W^{l}_{r}h_{u}^{l} \right ),
\end{equation*}
where $\mathcal{N}_{r}(v)$ is the set of all nodes $u$ connected to node $v$ via an edge with the label $r\in R$, and $W^{l}_{r}$ is a label-specific weight matrix. 
Each layer is followed by a Mish activation function~\cite{misra2020mish}.

\paragraph{Relational graph neural network.}
The \emph{relational graph neural network} (R-GNN) is a specialized GNN architecture that operates directly on the relational structure of states instead of translating them into graphs~\cite{staahlberg2022learning}.
Like \objectencoding and \objectatomencoding, the \rgnn also introduces goal predicates $P_{\mathcal{G}}$ to encode goal conditions into states.

\begin{algorithm}[h]
	\Input{Set of atoms $p \in s$ and objects $o \in s$}
	\Output{Object embeddings $h^{L}_{o}$}
	$h^{0}_{o} \gets \mathbf{0}^{k}$ for each object $o \in s$ \;
	\For{$l = 1, \dots, L$}{
		\For{\textbf{each} atom $p := P(o_{1}, ...,  o_{n}) \in s$}{
			$m_{p, o_{u}} \gets \left[\text{MLP}_{P}(h^{l}_{o_{1}}, ..., h^{l}_{o_{n}})\right]_{u}$ \;
			}
		\For{\textbf{each} object $o$}{
			$m_{o} \gets\text{aggregate}(\multiset{m_{p,o} \: \vert \: o \in p})$ \;
			$h^{l+1}_{o} \gets h^{l}_{o} + \text{MLP}_{U}(h^{l}_{o}, m_{o})$ \;
		}
	}
	\caption{\label{alg:rgnn}Relational graph neural network.}
\end{algorithm}

Consider Algorithm~\ref{alg:rgnn}, which shows a forward pass of the R-GNN.
The object embeddings $h^{0}_{o}$ initially contain only zeros.
The R-GNN then updates them over $L$ layers, each consisting of two steps:
\begin{enumerate}
	\item \emph{Message Computation} (lines 3-4): For each atom $p := P(o_{1}, ...,  o_{n}) \in s$, we pass the involved objects' embeddings $h^{l}_{o}$ to a predicate-specific MLP$_{P}$, which computes a message $m_{p, o}$ for each argument $o$ of $p$.
	\item \emph{Embedding Update} (lines 5-7): 
	For each object $o$, we aggregate the messages $m_{p,o}$ into a single message $m_{o}$ by applying a dimension-wise smooth maximum. 
	The updated embedding $h^{l+1}_{o}$ is then computed by passing $h^{l}_{o}$ and $m_{o}$ to MLP$_{U}$, with the addition of a residual connection.
\end{enumerate}

\paragraph{Computing state or Q-values.}
All three GNN architectures have in common that they iteratively compute embeddings for every object in a given state.
Independent of the architecture, these embeddings can then be used to predict either a single state-value or a Q-value for every applicable action.
We here follow the approach of \citeauthor{stahlberg2025first-order}~\shortcite{stahlberg2025first-order}.

First, we compute a dimension-wise sum over the (node) embeddings of all objects to obtain a single state embedding
\begin{equation*}
    h_{s} = \sum_{o \in \mathcal{O}}h^{L}_{o}.
\end{equation*}
To compute a state-value, we simply pass the state embedding to a final MLP
\begin{equation*}
    V(s) = \text{MLP}(h_{s}).
\end{equation*}
To compute Q-values, however, we need to encode the applicable actions $a \in Act(s)$ into each state $s$.
Accordingly, we introduce a predicate $P_{A}$ for each action scheme $A \in \mathcal{A}$ and action objects $o_{a}$ for all ground actions $a \in Act(s)$.
We then extend each state with new ground atoms $p:= P_{A}(o_{a},o_{0},\dots,o_{n})$.
The \rgnn, \objectencoding, and \objectatomencoding architectures then process the extended state encodings as they did before.
Finally, we compute the Q-values by concatenating each ground action embedding with a copy of the state embedding and passing it to an MLP
\begin{equation*}
    Q(s,a) = \text{MLP}(h^{L}_{a} \parallel h_{s}).
\end{equation*}

\section{Training Details}
\label{app:training}
We here provide additional details about our training setup.

\paragraph{Hyperparameters.}
Each policy is trained using three random seeds, where each training run consists of up to $100$ epochs with a batch size of $256$.
Policy updates are computed using the Adam optimizer~\cite{kingma2014adam} with a learning rate of $0.0002$ for the state-value and vanilla Q-value models, and a higher learning rate of $0.002$ for the regularized Q-value models, since we found them to be less prone to overfitting.
We additionally limit the magnitude of each policy update using gradient clipping with a value of $0.1$.
For the R-GNN architecture, we use a hidden size of $32$ and $30$ layers with shared parameters.
For the RGCN architecture, which processes the OE and OAE graphs, we use a hidden size of $32$ and $30$ layers with separate parameters.

\paragraph{Data sets.}
For a given domain, we check for each size $n \in \left[2,\dots,100\right]$ whether it is possible to generate a corresponding instance.
If so, we uniformly sample generator inputs that yield an instance of size $n$ and pass them to the generator.
The generators were either taken from the IPC'23~\cite{ayal:etal:aim-23} or from the FF domain collection\footnote{\url{https://fai.cs.uni-saarland.de/hoffmann/ff-domains.html}}.
If the generator returns an instance that has already been generated or where the initial state already satisfies the goal, we discard it.

After generating all instances, we pass each to the \textit{seq-opt-merge-and-shrink} configuration of Fast Downward~\cite{helmert:jair-06} with limits of $20$ minutes and $64$ GB, and delete every instance for which no optimal plan was found.
Table~\ref{tab:data_set_sizes} shows the sizes of the solved instances.
Then, for each optimal plan, we compute the $h^{\mathrm{LMcut}}$~\cite{helmert2009landmarks} values of the siblings of each state along the optimal trajectory.
The final data sets then consist of tuples $\left\langle s, h^{\ast}(s), a^{\ast}, a_{1}, \dots, a_{m}, h^{\mathrm{LMcut}}(s'_{1}), \dots, h^{\mathrm{LMcut}}(s'_{m})\right\rangle$, where $s$ is a state on the optimal trajectory, $h^{\ast}(s)$ is its optimal heuristic value, $a^{\ast}$ is the teacher action, $a_{i}$ are the non-teacher actions, and $h^{\mathrm{LMcut}}(s'_{i})$ are the heuristic values of the successors $s'_{i}$ reached through $a_{i}$.

\begin{table}[h]
  \centering
  \setlength{\tabcolsep}{5pt}    
  \renewcommand{\arraystretch}{1.25}
  \scalebox{1.0}{
    \begin{tabular}{l *{1}{c}}
        Domain & Training Set Sizes \\
        \midrule
		Blocksworld & $[2-16]$ \\

		Childsnack & $[8-38]$ \\

		Ferry & $[3-25]$ \\

		Floortile & $[7-18]$ \\

		Rovers & $[10-26]$ \\

		Satellite & $[5-31]$ \\

		Transport & $[6-38]$  \\

		Gripper & $[2-25]$ \\
        
        Logistics & $[6-40]$ \\

        Visitall & $[4,9,16,25,...,81]$ \\

    \end{tabular}
  }
  \caption{Number of objects in training instances.}
  \label{tab:data_set_sizes}
\end{table}

\end{document}